\documentclass{article}
\usepackage{spconf,amsmath,graphicx,hyperref}
\usepackage{cleveref}
\usepackage{booktabs}

\title{Adaptive Runge-Kutta Dynamics for Spatiotemporal Prediction}
\name{Xuanle Zhao\textsuperscript{\rm 1,2}, Yue Sun\textsuperscript{\rm 1}, Ziyi Wang\textsuperscript{\rm 1,2}, Bo Xu\textsuperscript{\rm 1,2}, Tielin Zhang\textsuperscript{\rm 1}}
\address{
\textsuperscript{\rm 1} The Key Laboratory of Cognition and Decision Intelligence for Complex Systems,\\ Institute of Automation, Chinese Academy of Sciences \\
\textsuperscript{\rm 2} School of Artificial Intelligence, University of Chinese Academy of Sciences, Beijing, China \\}
\begin{document}
\ninept
\maketitle
\begin{abstract}
Spatiotemporal prediction is important in solving natural problems and processing video frames, especially in weather forecasting and human action recognition. Recent advances attempt to incorporate prior physical knowledge into the deep learning framework to estimate the unknown governing partial differential equations (PDEs) in complex dynamics, which have shown promising results in spatiotemporal prediction tasks. However, previous approaches only restrict neural network architectures or loss functions to acquire physical or PDE features, which decreases the representative capacity of a neural network. Meanwhile, the updating process of the physical state cannot be effectively estimated. To solve the problems mentioned above, we introduce a physical-guided neural network, which utilizes 
an adaptive second-order Runge-Kutta method with physical constraints to model the physical states more precisely. Furthermore, we propose a frequency-enhanced Fourier module to strengthen the model's ability to estimate the spatiotemporal dynamics. We evaluate our model on both spatiotemporal and video prediction tasks. The experimental results show that our model outperforms several state-of-the-art methods and performs the best in several spatiotemporal scenarios with a much smaller parameter count.
\end{abstract}
\section{Introduction}
\label{sec:intro}
Data-driven spatiotemporal prediction is a pivotal research area in deep learning, with critical applications in fields such as traffic flow forecasting \cite{zhang2017deep} and weather prediction \cite{rasp2020weatherbench}. To model complex dynamics and improve predictive accuracy, prominent methods utilize architectures such as convolutional neural networks (CNNs) \cite{gao2022simvp}, recurrent neural networks (RNNs) \cite{chang2021mau}, and transformers \cite{liu2022video} to capture spatial and temporal features effectively. For example, SimVP \cite{gao2022simvp} utilize a modified UNet structure to directly learn input-output sequential mapping without explicitly modeling temporal dependencies. Transformer-based methods like the Video Swin Transformer \cite{liu2022video} and Poolformer \cite{yu2022metaformer} implicitly leverage the attention mechanisms to model temporal dependencies. 
In contrast to other approaches, recurrent-based methods are intuitively suited for spatiotemporal prediction due to their inherent strength in modeling sequential data. For example, the ConvLSTM \cite{shi2015convolutional} integrates convolutions into the recurrent cell. Building on this structure, subsequent works have proposed more sophisticated designs. For example, E3D-LSTM \cite{wang2018eidetic} and CrevNet \cite{yu2020efficient} incorporate complex 3D convolutions, while the PredRNN series \cite{wang2017predrnn} introduce modified internal memory structures to capture long-range dependencies.

While powerful, data-driven approaches struggle in scenarios with scarce or noisy data, producing physically inconsistent predictions. To mitigate this, physics-informed methods such as PINNs \cite{raissi2019physics} and DeepONets \cite{lu2021learning} embed physical laws as inductive biases, which proves effective in domains where such governing equations are known. However, this reliance on explicit, domain-specific priors makes them overly specialized and difficult to generalize to problems where the underlying dynamics are complex or unknown \cite{kapoor2023physics}. In the spatiotemporal prediction field, previous works have utilized prior physical knowledge in neural networks to improve the neural networks with more physical insights and shown promising improvements in spatiotemporal prediction. For example, PhyDNet \cite{guen2020disentangling} regularizes its kernel weights by incorporating physical laws into the loss function. This encourages the model to learn features corresponding to partial derivatives and the underlying physical dynamics of the prediction task.
Many similar works \cite{guen2020disentangling} have also been proposed to solve tasks in specific domains, such as precipitation \cite{zhang2023skilful} and weather forecasting \cite{verma2023climode}.

This motivates a hybrid approach where physical priors, whether architectural or equation-based, are integrated with data-driven methods. To this end, we propose a novel method that improves spatiotemporal prediction accuracy by integrating data-driven and physics-informed approaches. Specifically, we propose a novel physics-guided recurrent neural network composed of a parallel pipeline that enhances spatial representations with Fourier blocks, and a physics-guided adaptive second-order Runge-Kutta module to estimate and update the underlying physical state. Besides, we introduce the H1 loss to enhance the high-frequency information.
The contributions of this work are summarized as follows: 
(1) We propose a dual-pipeline architecture that learns robust spatiotemporal representations by combining frequency and spatial domains. (2) We propose an Adaptive Runge-Kutta Module (ARKM) that integrates a second-order update with an adaptive gating mechanism.
(3) We introduce high-frequency H1 loss for high-frequency learning and integrate it with the MSE loss and moment loss to collectively capture fine-grained details. 
(4) Our model achieves superior performance across a wide range of benchmarks, while being significantly more parameter-efficient than previous methods.

\begin{figure*}[t]
    \centering
    \includegraphics[width=0.98\linewidth]{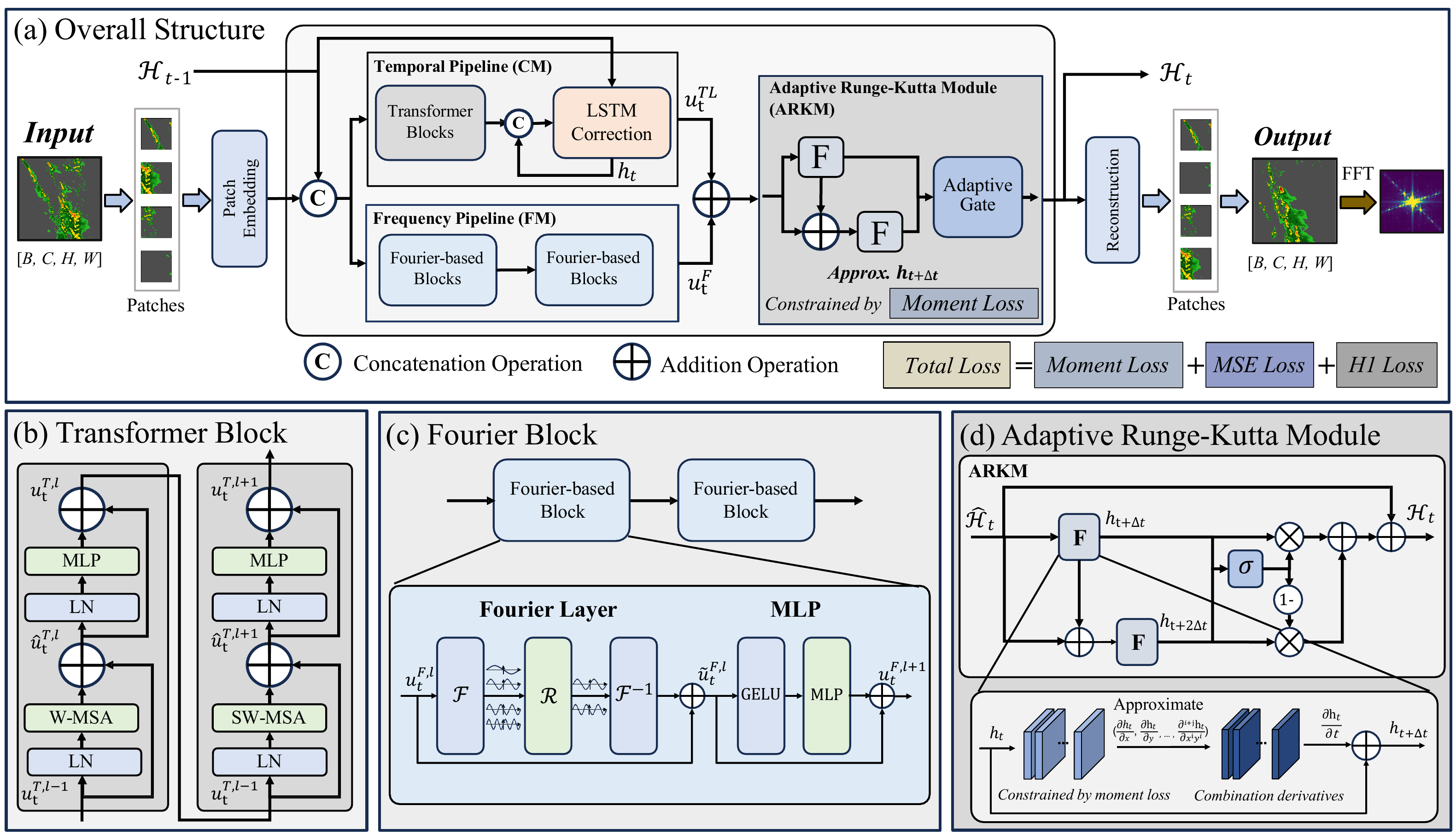}
    \vspace{-10pt}
    \caption{The overall and detailed structures of our proposed model. (a). The overall network architecture. The total loss contains the moment, MSE, and H1 loss. (b) The detailed structure of Transformer Blocks. (c) The detailed structure of Fourier Blocks. (d) The detailed structure of the Adaptive Runge-Kutta Module. The function $F$ in this module consists of convolution networks with moment loss constraints and another convolution for combination.}
    \label{fig:model}
    \vspace{-10pt}
\end{figure*}

\section{Related Works}
Spatiotemporal prediction methods are generally divided into recurrent-based and recurrent-free approaches. Recurrent-based models historically received more attention. ConvLSTM \cite{shi2015convolutional} pioneered the field by incorporating convolutions into LSTM cells, a design later extended by models like PredRNN \cite{wang2017predrnn}. More recent approaches have integrated stronger priors; for example, PhyDNet \cite{guen2020disentangling} explicitly disentangles physical and residual dynamics, while SwinLSTM \cite{tang2023swinlstm} incorporates Transformer blocks within the recurrent cell. While these methods have achieved strong performance, many are purely data-driven. Recurrent-free methods include CNN-based models like SimVP \cite{gao2022simvp}, which directly map input to output frames. Transformer-based models like Video Swin Transformer \cite{liu2022video} that use attention to capture long-range dependencies. A promising recent direction is physics-based methods, particularly those operating in the frequency domain. The Fourier Neural Operator (FNO) \cite{li2020fourier} demonstrated the effectiveness of learning solution operators in Fourier space for solving PDEs, inspiring subsequent works \cite{wu2023earthfarseer} in dynamic systems simulation. 

Integrating physical knowledge, like partial differential equations (PDEs), into deep learning frameworks is a key research area. The approaches can be broadly grouped. One method is to embed known PDEs as explicit priors. This is common in domains with well-understood physics, like weather forecasting, where models such as ClimODE \cite{verma2023climode} and NowcastNet \cite{zhang2023skilful} directly incorporate convection equations into their architecture. When facing unknown governing equations, models can learn the underlying PDEs from data. Methods like PDE-Net \cite{long2018pde} and PhyDNet \cite{guen2020disentangling} use constrained convolutions to approximate differential operators, effectively discovering the dynamics in a data-driven way.

\section{Method}

\subsection{Model Architecture}\label{method:model}
As shown in \Cref{fig:model}, our model first embeds the input frame $x_t$ into a latent representation $u_t$ via patch embedding. 
This representation is then fed into two parallel branches: one using Transformer blocks for spatial feature extraction and another using LSTM cells for temporal dynamics. These branches are augmented with Fourier blocks to enhance spatial dependencies in a physically-guided way. An adaptive Runge-Kutta module then integrates the outputs. 

\textbf{Correction Module (CM).}
To efficiently model spatial features while avoiding the quadratic computational cost of the global multi-head self-attention (MHSA) used in Vision Transformers, our model's spatial encoder is built with Swin Transformer blocks \cite{liu2021swin}. These blocks replace global attention with more efficient window-based (W-MHSA) and shifted-window-based (SW-MHSA) mechanisms. While Swin Transformer blocks are powerful for capturing spatial correlations, they do not inherently model temporal dependencies. To address this, we incorporate an LSTM correction module. The LSTM explicitly models temporal coherence by integrating information from previous timesteps to update and refine the hidden features. After the temporal coherence, the hidden states are corrected according to the input feature of the current time.
\begin{equation}
\begin{aligned}
u_t^{CM}&=\left(1-\mathcal{K}_t\right) \odot\mathcal{H}_{t-1}+\mathcal{K}_t \odot u_t^{TL} \\
&=\mathcal{H}_{t-1}+\mathcal{K}_t \odot\left(u_t^{TL}-\mathcal{H}_{t-1}\right) 
\end{aligned}
\end{equation}
where $u_t^{TL}$ and $\mathcal{K}_t=\sigma(u_t^{TL})$ are the representations after the Swin Transformer and LSTM before, and the gating factor to balance the predicted and previous hidden state.

\textbf{Frequency Module (FM).} This module enhances the features from the Correction Module using a physics-informed data-driven approach. The core of the FM is the Fourier block, which performs a 2D Fourier Transform (FFT) on the input tokens, applies learnable kernels in the frequency domain, and then transforms the result back using an Inverse FFT (IFFT). This allows for direct modeling of the mapping function within the Fourier domain.
\begin{equation}
\mathcal{Z}^l_t(u, v)=\sum_{x=0}^{h-1} \sum_{y=0}^{w-1} u^l_t(x, y) e^{-2 \pi i\left(\frac{u}{h} x+\frac{v}{w} y\right)}.
\end{equation}
where $u^l_t(x, y)$ denotes the input of the $l$th Fourier block.
Then, a kernel $\mathcal{R}_\phi$ parameterized by $\phi$ with real and imaginary parts is trained to multiply with $\mathcal{Z}^l_t(u, v)$ to compute in the Fourier domain. 
\begin{equation}
\tilde{\mathcal{Z}}^l_t(u, v)=\left(\mathcal{R}^l_\phi\odot \mathcal{Z}^l_t(u, v)\right)
\end{equation}
Lastly, the mixed tokens are transformed from the Fourier domain to the spatial domain with 2D IFFT $\mathcal{F}^{-1}$. 
\begin{equation}
\begin{aligned}
\tilde{u}^l_t(x, y) & =\frac{1}{h w} \sum_{u=0}^{h-1} \sum_{v=0}^{w-1} \tilde{\mathcal{Z}}^l_t(u, v) e^{2 \pi i\left(\frac{u}{h} x+\frac{v}{w} y\right)} \\
\end{aligned}
\end{equation}
The detailed structure is shown in \Cref{fig:model} (c).

\textbf{Adaptive Runge-Kutta Module (ARKM).}
To update the hidden state from the Correction and Frequency Modules in a PDE-guided manner, we employ an adaptive second-order Runge-Kutta (ARK2) method. While the commonly used first-order Euler method can be viewed as a simple residual connection, ARK2 provides a more accurate numerical integration of the system's dynamics. We first introduce the conventional RK2 method as follows.
\begin{equation}\label{eq:rk2}
\begin{aligned}
&h_{t+1} = h_t+\frac{1}{2}\left(h_{t+\Delta t}+h_{t+2\Delta t}\right), \\
&h_{t+\Delta t} =F_\theta\left(h_t\right), h_{t+2\Delta t} =F_\theta\left(h_t+h_{t+\Delta t} \right),
\end{aligned}
\end{equation}
where $h_{t+\Delta t}, h_{t+2\Delta t}$ are assumed as the approximate solution at intermediate steps $t+\Delta t$ and $t+2\Delta t$ respectively and $\Delta t=1/3$. Furthermore, the temporal derivative could be computed by combining spatial derivatives $F_s$. 
\begin{equation}\label{eq:parial}
\begin{aligned}
\frac{\partial h_t}{\partial t}  =F_s\left(\frac{\partial h_t}{\partial x}, \frac{\partial h_t}{\partial y}, \frac{\partial^2 h_t}{\partial x \partial y},  \ldots, \frac{\partial^{i+j} h_t}{\partial x^i \partial y^j}\right) \\
\end{aligned}
\end{equation}
While standard convolutions can approximate spatial derivatives, they are not guaranteed to learn the correct finite difference operators. To address this, we introduce a moment loss to explicitly constrain the convolutional kernels, forcing them to compute partial derivatives accurately.

Specifically, the procedure could be separated as utilizing a moment loss constrained convolution network $\operatorname{Conv2d}_{k\times k}$ with $k^2$ kernels of size $k\times k$ to approximate the spatial derivatives $(\frac{\partial h_t}{\partial x}, \frac{\partial h_t}{\partial y}, \frac{\partial^2 h_t}{\partial x \partial y}, \ldots, \frac{\partial^{i+j} h_t}{\partial x^i \partial y^j})$. Then another convolution network $\operatorname{Conv2d}_{1\times 1}$ with $1\times 1$ kernel size combines the spatial derivative into temporal derivatives. In this way, the temporal derivatives are precisely estimated.

We observe that the conventional RK2 method is susceptible to gradient vanishing as network depth increases. To solve this, we propose an adaptive RK2 method to scale the $h_{t+\Delta t}$ and $h_{t+2\Delta t}$ by introducing learnable gate coefficients.
\begin{equation}
\begin{aligned}
\mathcal{H}_{t} & =\hat{\mathcal{H}}_{t}+g \cdot h_{t+\Delta t}+(1-g) \cdot h_{t+2\Delta t} \\
g & =\operatorname{sigmoid}\left(W_g * \left(h_{t+\Delta t}; h_{t+2\Delta t}\right)+b_g\right),
\end{aligned}
\end{equation}
where the adaptive gate is computed by $1\times1$ kernel convolution $W_g$ and bias $b_g$. $\hat{\mathcal{H}}_{t}$ is the sum of $u_t^{F}$ and $u_t^{CM}$. The architecture of the detailed structure is shown in \Cref{fig:model} (d). Following the conventional RK2 method, we employ a single function $F_\theta$ with shared parameters to compute intermediate steps $h_{t+\Delta t}$ and $h_{t+2\Delta t}$.

\subsection{Loss Fucntions}\label{subsec:loss}
In our experiments, we minimize the following loss to optimize our model:
$\mathcal{L} = \mathcal{L}_{\text {prediction}}+\lambda_m\mathcal{L}_{\text {moment}}$.
The prediction loss $\mathcal{L}_{\text {prediction}}$ contains two losses, the MS and H1 loss, which could be denoted as  $\mathcal{L}^{MSE}+\lambda_H\mathcal{L}^{H1}$.
In visual tasks, critical information is contained within high-contrast pixels, corresponding to high-frequency components in the frequency domain. To learn the high-frequency feature precisely, we introduce the frequency domain H1 loss to emphasize the high-frequency components in the image.
\begin{equation}
\mathcal{L}^{H^1}(\boldsymbol{\hat{x}}, \boldsymbol{y})=\sum\left(1+4 \pi^2|\xi|^2\right)\left|\mathcal{F}(\boldsymbol{\hat{x}})_{\xi}-\mathcal{F}({\boldsymbol{y}})_{\xi}\right|^2.
\end{equation}
The H1 loss introduces a weighting term proportional to $|\xi|^2$, where $\xi$ is the frequency. This effectively makes it a weighted L2 loss that penalizes errors in high-frequency components.

\begin{table*}[t]
	\centering
        \small
        \caption{Quantitative comparison of our method and other methods on \textbf{Moving MNIST}, \textbf{TaxiBJ} and \textbf{KTH}. Parameters are counted when training the \textbf{Moving MNIST}. The \textbf{Bold} and \underline{Underline} denote the best and second-best performance, respectively.}
        $
	\begin{array}{c|c|ccc|cc|ccc} 
            \toprule 
            & \text{Parameters} & \multicolumn{3}{c}{\text{Moving MNIST}} &  \multicolumn{2}{c}{\text{TaxiBJ}} &  \multicolumn{3}{c}{\text{KTH}} \\
            \text { Method } & \text{(M)} & \text { MSE } \downarrow &  \text { MAE } \downarrow &\text { SSIM }\uparrow  & \text { MAE } \downarrow & \text { SSIM }\uparrow &\text { MSE } \downarrow &\text { MAE } \downarrow & \text { SSIM }\uparrow\\
		\midrule 
		\text { ConvLSTM~\cite{shi2015convolutional} }  & \text{15.0}& \text{29.80} & \text{90.64} &\text{0.9288} & \text{15.32} & \text{0.9836} & \text{47.65} & \text{445.50} & \text{0.8977}\\ 
		\text { PredRNN~\cite{wang2017predrnn} } & \text{38.6} & \textbf{24.53} &  \textbf{73.12}& \textbf{0.9462} & \text{15.37} & \text{0.9834} & \text{41.07} & \text{380.60} & \underline{0.9097}\\
		\text { E3D-LSTM~\cite{wang2018eidetic} } &\text{51.0} & \text{35.97} & \text{78.28} &\text{0.9320} & \text{15.26} & \text{0.9834} & \text{136.40} & \text{892.70} & \text{0.8153} \\
		\text { MAU~\cite{chang2021mau} }  & \text{4.5} &\text{26.86} & \text{78.22} &\text{0.9398} & \text{15.26} & \text{0.9834} & \text{51.02} & \text{471.20} & \text{0.8945}\\
            \text {CrevNet~\cite{yu2020efficient}} &\text{5.0} &\text{30.15} & \text{86.28} &\text{0.9350} & \text{15.73} &\text{0.9821} &\text{61.32} & \text{503.15} &\text{0.8815} \\
		\text { PhyDNet~\cite{guen2020disentangling}}  & \text{3.1} &\text{28.19} &\text{78.64} &\text{0.9374} & \text{15.53} & \text{0.9828} & \text{91.12} & \text{765.60} & \text{0.8322}\\
		\text {SwinLSTM~\cite{tang2023swinlstm} } & \text{20.1} &\text{27.74} & \text{77.21} &\text{0.9313} &\text{15.74} & \text{0.9813}&\text{45.23} &\text{437.51} & \text{0.9030} \\
            \text {DMVFN~\cite{hu2023dynamic} } &\text{3.5} &\text{123.67} & \text{179.96} & \text{0.8140} &\text{15.72} & \text{0.9833} &\text{59.61} &\text{413.20} & \text{0.8976}\\   
            \text {SimVP~\cite{gao2022simvp} } & \text{58.0}&\text{32.15} & \text{89.05} & \text{0.9268} &\text{15.45} & \text{0.9835}  &\text{41.11} &\text{397.10} & \text{0.9065}\\ 
            \midrule
            \text{Ours} & \text{3.8} &\underline{24.83} &\underline{74.83} & \underline{0.9423} &\textbf{14.88} & \textbf{0.9851} &\textbf{39.27} & \textbf{370.56} & \textbf{0.9173}\\
            \text{Ours(w/o H1)} & \text{3.8} & \text{25.19} &\text{75.74} & \text{0.9401}  &\underline{14.96} & \underline{0.9846} &  \underline{40.02} &\underline{376.25} & \text{0.9088}\\ 
		\bottomrule
	\end{array}
        \label{table:video}    
        $
        \vspace{-15pt}
\end{table*}

\begin{table}[t]
	\centering
        \small
        \caption{Quantitative comparison of our method and other methods on \textbf{SEVIR}, \textbf{Navier-Stokes} and \textbf{Weather} temperature benchmarks. * indicates that our implementation. }
        $ 
	\begin{array}{c|c|c|c} 
            \toprule 
              & \multicolumn{1}{c}{\text{SEVIR}} &  \multicolumn{1}{c}{\text{Navier-Stokes}} &  \multicolumn{1}{c}{\text{Weather}} \\
            \text { Method }  & \text {CSI-M}\times \text {100}\uparrow  & \text {N-MSE} \times \text {100} \downarrow & \text { MSE }\downarrow \\
		\midrule 
		\text { ConvLSTM} & \text{36.73}   & \text{23.87}   & \text{1.521}\\ 
		\text { PredRNN }  & \text{41.13}  & \text{19.53}   & \text{1.331}\\
            \text{FNO} & \text{-}  & \text{19.82}  & \text{1.491}\\
		\text { MAU } & \text{44.75}  & \text{18.43}  & \text{1.251}\\
            \text {SimVP}  & \text{45.84}  & \text{15.98}& \text{1.238} \\ 
             \text{Rainformer}  & \text{38.72}  & \text{18.31}  & \text{-} \\ 
          \text{EarthFormer*}   & \text{44.73}  & \underline{15.39}  & \text{-} \\ 
        \text{Earthfarsser*}   & \underline{46.54}  & \text{15.64} & \underline{1.121} \\ 

            \midrule
            \text {Ours}  & \textbf{47.02}  & \textbf{14.89}  & \textbf{1.086} \\
		\bottomrule
	\end{array}
        \label{table:natural}
        $
        \vspace{-15pt}    
\end{table}

\begin{table}[t]
	\centering
        \small
        \caption{Ablation study on different patch sizes.}
        $
	\begin{array}{c|cc|cc} 
            \toprule 
             & \multicolumn{2}{c}{\text{TaxiBJ}}  & \multicolumn{2}{c}{\text {KTH}}  \\
            \text { Patch size }  & \text{MAE} \downarrow & \text{SSIM} \uparrow   & \text{MSE} \downarrow & \text{SSIM} \uparrow\\ 
		\midrule 
		\text { 2 }  & \text{15.52} & \text{0.9839}  & \text{43.16} & \text{0.9005} \\ 
		\text { 4 }  & \textbf{14.88} & \textbf{0.9851}  & \textbf{40.17} & \textbf{0.9143}\\  
		\text { 8 }  & \text{16.02} & \text{0.9818}  & \text{53.98} & \text{0.8916} \\ 
		\bottomrule
	\end{array}
        $
        \label{table:abpatch}
        \vspace{-10pt}
\end{table}
To force the convolution network in the Adaptive RK2 Module to approximate the spatial derivatives, we follow PhyDNet \cite{guen2020disentangling} to impose the moment loss $\mathcal{L}_{\text {moment}}$ to constrain the weights of the convolutions. Specifically, assuming a convolution network is parameterized by $\mathbf{w}_p=\left\{\mathbf{w}_{p, i, j}^k\right\}_{i, j \leq k}$ with $k^2$ filter of size $k\times k$. We aim to enforce each filter $\mathbf{w}_{p, i, j}^k$ to approximate the spatial derivative of order $(i, j): \frac{\partial^{i+j} h}{\partial x^i \partial y^j}$ by using the moment loss. The calculation of the moment loss is derived from the filter as $\mathcal{M}\left(\mathbf{w}_{p, i, j}^k\right)$ and compared to the target moment matrix $M_{i, j}^k$ via:
\begin{equation}\label{loss:moment}
\mathcal{L}_{\operatorname{moment}}\left(\mathrm{w}_p\right)=\sum_{i \leq k} \sum_{j \leq k} \mathcal{L}^2\left(\mathcal{M}\left(\mathrm{w}_{p, i, j}^k\right), M_{i, j}^k\right)
\end{equation}
where $\mathcal{L}^2$ is the squared error.
\section{Experiments}
\label{sec:experiments}
We evaluate our proposed model on various benchmarks, including synthetic, human video, and natural scene datasets.

\subsection{Main Results}
\textbf{Moving MNIST \cite{srivastava2015unsupervised}.} This is a standard benchmark for video prediction. Each sequence features two handwritten digits moving and bouncing within a $64\times64$ frame. Following the OpenSTL \cite{tan2024openstl}, we predict the last 10 frames given the initial 10 as input.

\textbf{TaxiBJ \cite{zhang2017deep} .} TaxiBJ is a real-world traffic benchmark from Beijing, consisting of $32\times32\times2$ maps with two channels for traffic inflow and outflow, recorded at 30-minute intervals. The task is to predict future traffic patterns from historical sequences.

\textbf{KTH \cite{schuldt2004recognizing}.} KTH is a classic benchmark for human action prediction. It shows 25 people doing six different actions (like walking, boxing, etc.) in four scenarios. We use a $128\times128$ resolution and task our model with predicting the next 20 video frames.

\textbf{SEVIR \cite{veillette2020sevir}.} SEVIR contains weather radar data, specifically Vertically Integrated Liquid (VIL), with a frame resolution of $384\times384$ and a 5 minute interval between frames. For our experiments, we follow the preprocessing steps from EarthFormer \cite{gao2022earthformer} and use the 2017 portion of the dataset for training and evaluation.

\textbf{Navier-Stokes \cite{li2020fourier}..} The Navier-Stokes equation on a two-dimensional torus is a standard benchmark for physics-informed machine learning. Following the baseline established in that work, we use the Normalized Mean Squared Error (N-MSE) for evaluation. 

\textbf{Weather \cite{rasp2020weatherbench}.} The Weather dataset is a large-scale climate prediction benchmark containing diverse variables. The data is downsampled to a $32\times64 $ grid (5.625° resolution) with a 1-hour time interval between frames. We use data from 2010-2015 for training, 2016 for validation, and 2017-2018 for testing.

Results in \Cref{table:video} show that our model performs best in the TaxiBj and KTH datasets and is second best in the Moving MNIST dataset. Notably, the number of parameters is much smaller than previous SOTA methods 
PredRNN \cite{wang2017predrnn}, SimVP \cite{gao2022simvp} and SwinLSTM \cite{tang2023swinlstm}. Also, as shown in \Cref{table:natural}, our model performs best in all the natural dynamic phenomena. Our model achieves the lowest MSE, and the highest CSI-M across these datasets.

\subsection{Ablation study}
To verify the impact of each component, we conduct ablation studies under different experimental conditions. 

\textbf{Patch size.} The path embedding size directly influences the number of model parameters and the length of the input token sequence. We conduct our ablation experiments on TaxiBj and KTH with patch size set to $2\times2$, $4\times4$, and $8\times8$. \Cref{table:abpatch} shows that the size of $4\times4$ generally works better.

\begin{table}[t]
	\centering
        \small
        \caption{Ablation study on different upsampling methods in the Decoder layer. \text{ConvT2D} and \text {Bi-Interp} are the abbreviations of \text{ConvTransposed2D} and \text{Bilinear Interpolation}.}
        $
	\begin{array}{c|cccc} 
            \toprule 
             & \multicolumn{2}{c}{\text{TaxiBJ}}  & \multicolumn{2}{c}{\text {KTH}}  \\
            \text {Method}  & \text{MAE} \downarrow & \text{SSIM} \uparrow   & \text{MSE} \downarrow & \text{SSIM} \uparrow\\ 
		\midrule 
		\text { ConvT2D }  & \textbf{14.88} & \textbf{0.9851}  & \textbf{40.17} & \textbf{0.9143} \\ 
		\text {Bi-Interp}  & \text{16.43} & \text{0.9637}  & \text{50.36} & \text{0.8643}\\
		\bottomrule
	\end{array}
        $
        \label{table:abup}
\end{table}

\textbf{The Decoder Layer.} We design the experiments on two upsampling methods, ConvTransposed2D and Bilinear Interpolation, which are two commonly used vision tasks. \Cref{table:abup} shows that the ConvTransposed2D is much better than Bilinear Interpolation.

\begin{figure}[t]
    \centering
    \includegraphics[width=0.95\linewidth]{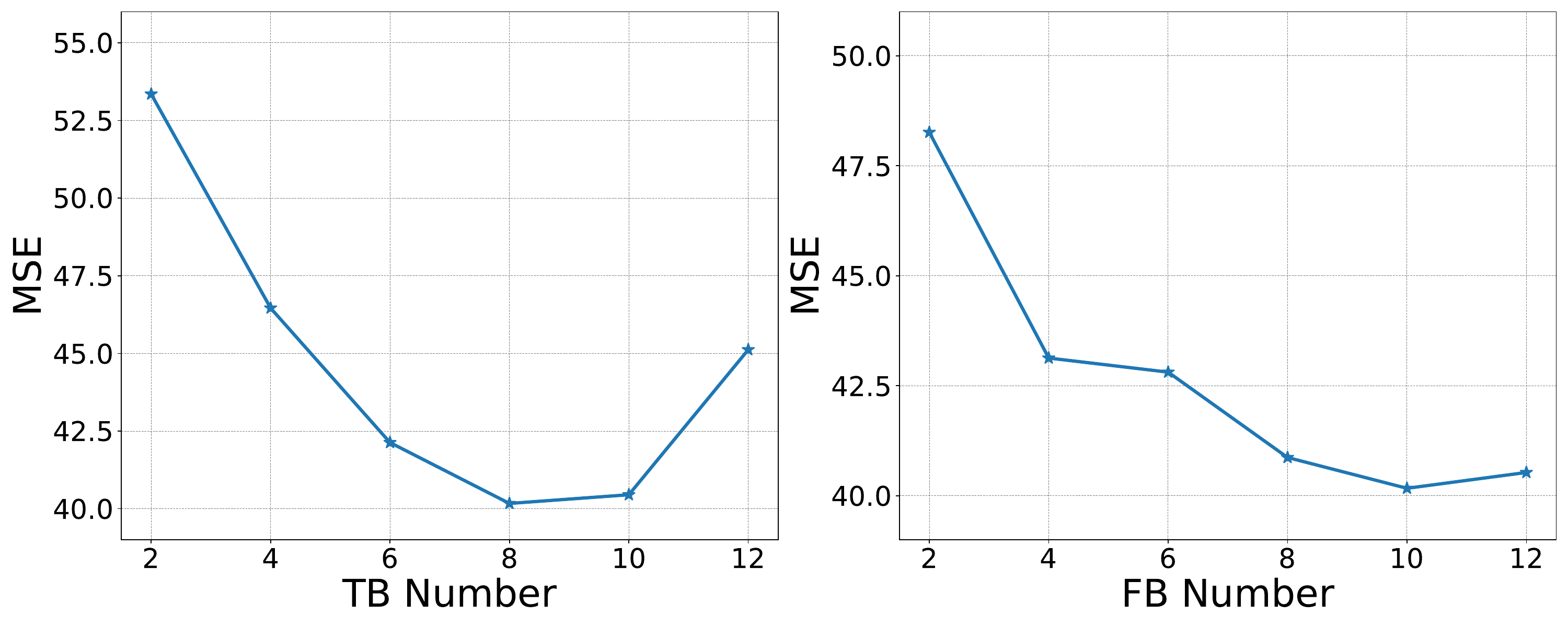}
    \vspace{-10pt}
    \caption{Ablation study on the various numbers of Transformer Blocks (TB) and Fourier Blocks (FB) on the KTH dataset.}
    \vspace{-10pt}
    \label{fig:ab}
\end{figure}

\textbf{Number of blocks.} The number of blocks in each module highly influences the learning capacity of the model and the number of parameters. Owing to the limitation of computation resources, we fix the number of Transformer blocks to 8 or the number of Fourier blocks to 10 (the best setting in the KTH dataset) to explore the influence of the other blocks. The results are shown in \Cref{fig:ab}.

\section{Conclusion}
This paper introduces a novel method that integrates Fourier blocks with Transformers to capture high-frequency features. The dynamics are governed by a new adaptive Runge-Kutta (RK2) module, which uses physics-guided convolutions to estimate and update the hidden state. To ensure the model learns both the underlying physics and fine visual details, our training objective combines a moment loss, an H1 loss, and a standard MSE loss. Our result shows that our model performs better in spatiotemporal and video prediction tasks with fewer parameters.

\small{
\bibliographystyle{IEEEbib}
\bibliography{refs}
}
\end{document}